\documentclass{article}

\usepackage{multirow}
\usepackage{graphicx}

    \usepackage[final]{timeseries_workshop}

\usepackage[utf8]{inputenc} 
\usepackage[T1]{fontenc}    
\usepackage{caption}        
\usepackage{hyperref}       
\usepackage{url}            
\usepackage{booktabs}       
\usepackage{amsfonts}       
\usepackage{nicefrac}       
\usepackage{microtype}      
\usepackage{xcolor}         
\usepackage{svg}
\usepackage{wrapfig}

\title{Joint Embeddings Go Temporal}

\author{%
  Sofiane Ennadir\thanks{Work done during internship at Flatiron Institute - Corresponding Author: \texttt{ennadir@kth.se}} \\
  KTH \\
  Stockholm, Sweden\\
  \And
  Siavash Golkar  \\
  New York University\\
  New York, USA \\
  \And
  Leopoldo Sarra \\
  Flatiron Institute \\
  New York, USA \\
}

\begin{document}

\maketitle

\begin{abstract}
  Self-supervised learning has seen great success recently in unsupervised representation learning, enabling breakthroughs in natural language and image processing. However, these methods often rely on autoregressive and masked modeling, which aim to reproduce masked information in the input, which can be vulnerable to the presence of noise or confounding variables. To address this problem, Joint-Embedding Predictive Architectures (JEPA) has been introduced with the aim to perform self-supervised learning in the latent space. To leverage these advancements in the domain of time series, we introduce Time Series JEPA (TS-JEPA), an architecture specifically adapted for time series representation learning.  We validate TS-JEPA on both classification and forecasting, showing that it can match or surpass current state-of-the-art baselines on different standard datasets. Notably, our approach demonstrates a strong performance balance across diverse tasks, indicating its potential as a robust foundation for learning general representations. Thus, this work lays the groundwork for developing future time series foundation models based on Joint Embedding.

\end{abstract}

\section{Introduction}
The paradigm of self-supervised learning (SSL) has emerged as a crucial technique for the development of Foundation Models~\cite{bommasani2021opportunities}. These models are built on large unlabeled datasets and then fine-tuned for specific tasks with smaller labeled datasets. This approach has been very successful for language and vision tasks, and has been recently applied also to time series~\cite{das2023decoder, goswami2024moment}.
SSL methodologies can be broadly categorized into two families.
The first is based on contrastive approaches~\cite{zhang2022survey}, such as SimCLR~\cite{chen2020simple} developed for computer vision, which learns representations by juxtaposing positive and negative pairs of samples.
Building upon these principles, researchers have also introduced novel adaptations~\cite{yue2022ts2vec, zhang2022survey, zhang2022self} that account for the specific properties of time series.
The second family encompasses predictive or generative-based methods. These techniques, such as the Masked Auto-Encoder (MAE)~\cite{he2022masked}, task the model to reconstruct or predict missing segments of the input.
This approach can be particularly effective for time series, given their inherent temporal ordering and often cyclical structure, such as in seasonal retail sales \cite{ensafi2022time}.
A variation of MAE is the Autoregressive Encoder, where the model's objective becomes the prediction of the next time steps. This adaptation aligns particularly well with the forward-looking nature of many time series applications, enhancing the capacity to capture temporal dependencies specifically in the forecasting task.

However, since masked models are designed to reconstruct missing parts in the input space, they can be susceptible to the presence of noise and other non-predictable confounding factors. 
Indeed, a good reconstruction requires to model the entire input, and such elements may prevent the extraction of meaningful and predictive features, because the model will focus on this noise rather than the underlying patterns within the data~\cite{lecun2022path}.
Recently, the Joint-Embedding Predictive Architecture (JEPA)~\cite{lecun2022path} has been introduced as a way to only maintain the relevant information in the representation, leading to competitive results in both the image~\cite{assran2023self} and video domains~\cite{bardes2024vjepa}.
The JEPA approach first encodes the input into a latent space and then performs masked reconstruction in this space. Latent space reconstruction makes this technique robust to the presence of confounding variables and noise in the input. 
Inspired by the success of JEPA in addressing confounding factors in images and videos, we believe that this approach can similarly benefit time series data, which are often inherently noisy.

Here, we propose TS-JEPA, an adaptation of the JEPA architecture  specifically tailored to the unique characteristics of temporal sequences.
We validate our model through experiments on various standard datasets for time series classification and short and long-term forecasting. TS-JEPA exhibits promising performance compared to other baselines with the induced representation performing competitively in the forecasting task while outperforming significantly on classification.
While a few papers, have already appeared mentioning JEPA in the context of time series, they were either a specific application to encoded frames~\cite{girgis2024time}, or in combination with other techniques for in-context prediction~\cite{verdenius2024lat}.
Ours is the first systematic study of the JEPA architecture to time series, investigating whether this method can compete against the more common autoregressive approach.
Our results positions TS-JEPA as a promising building block to be used for time series foundation models.

\section{Joint Embedding Predictive Architecture for Time Series}
We consider the task of representation learning for time series.
We focus on the univariate case, but our architecture is easily adaptable to multivariate time series.
Formally, let $\mathcal{X} = \{x_1, x_2, ..., x_N\}$ represent a set of $N$ time series, where each $x_j$ is of length $T$.
We aim to find a function $f: \mathcal{X} \rightarrow \mathcal{H}$ mapping the input time series to a representation $h \in \mathcal{H} \subset \mathbb{R}^m$, where $m$ is the hidden representation's dimension. Once the self-supervised representation learning process has been completed, the learned representations can be used for downstream tasks such as classification or forecasting.

The fundamental idea of JEPA  is to use non-masked parts of the input to predict other masked parts, but in an \emph{ abstract latent space} rather than in the original input space.
This approach allows the model to capture underlying patterns and relationships in the data that may not be immediately apparent in the raw input.
Our architecture is composed of four primary components~(Fig.~\ref{fig:architecture}):

\begin{figure}
  \centering
  \includegraphics[width=0.95\linewidth]{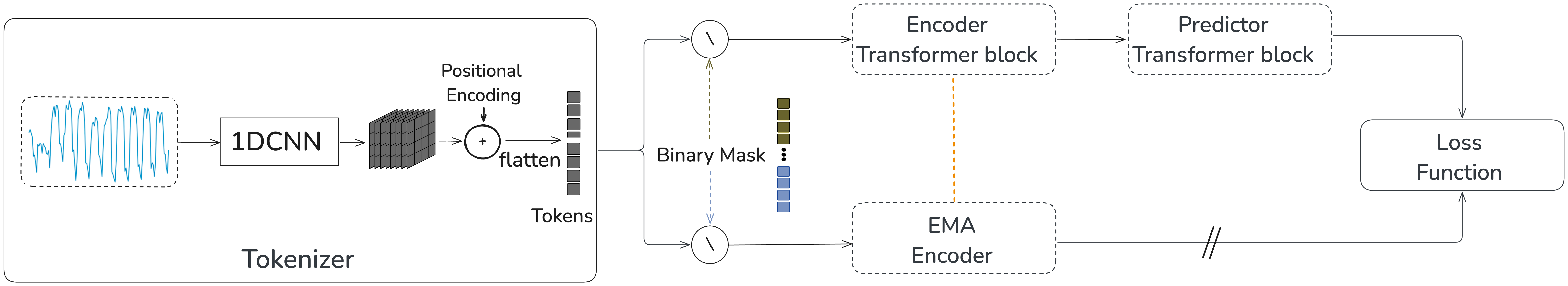}
  \caption{Illustration of TS-JEPA: it consists of ($1$) a tokenizer, ($2$) an encoder that processes the non-masked patches, ($3$) a Predictor that generates the target predictions from the encoder's output and ($4$) the EMA-Encoder, which encodes the target masked patches.}
  \label{fig:architecture}
  \vspace{-10pt}
\end{figure}

\hspace{1em} \textbf{(i) Tokenizer:} serves as the initial processing stage. It takes as input a time series $x \in \mathcal{X}$ and transforms it into a sequence of non-overlapping patches ${p_i \in \mathcal{P}}$.
To embed these patches, we use a one-dimensional convolutional neural network (1D-CNN).
The aim is to capture local patterns and features along the temporal dimension within each patch.
In addition, to preserve the temporal information, we incorporate a positional encoding mechanism.
While alternative positional encoding schemes could easily be incorporated in our architecture, we have adopted an absolute sin-cos positional embedding~\cite{vaswani2017attention} for simplicity.
The sequence of patches are split into a set of masked $\mathcal{P}_{\mathcal{M}}$ and non-masked patches $\mathcal{P}_{\mathcal{N}}$ following a uniform masking strategy.

\hspace{1em} \textbf{(ii) Encoder $E_{\theta}$:} is the cornerstone of our architecture, transforming the patches into a useful representation that can be used afterwards for  downstream tasks. We  use a standard transformer architecture incorporating self-attention mechanisms.
The encoder takes as input the non-masked patches $\mathcal{P}_{\mathcal{N}} = \{p_i \in \mathcal{P} \mid i \in \mathcal{N} \}$, where $\mathcal{N}$ represents the indices of non-masked elements. The inputs are transformed into the latents $z_{\mathcal{N}} = E_{\theta}(\mathcal{P}_{\mathcal{N}})$.

\hspace{1em} \textbf{(iii) Predictor $P_{\beta}$:} takes the output of the encoder $z_{\mathcal{N}}$ as  input.
The goal is to learn a mapping from the encoded observed (i.e. non-masked) tokens to the encoded unobserved (masked) ones.
Similar to the Encoder, we use a transformer-based architecture with self-attention mechanisms. Specifically, the output of the predictor can be formulated as $z'_{\mathcal{M}} = P_{\beta}(E_{\theta}(\mathcal{P}_{\mathcal{N}}))$.

\hspace{1em} \textbf{(iv) EMA-Encoder $E_{\bar{\theta}}$:} The Exponential Moving Average encoder is an instance of the encoder with different weights. Instead of them being directly optimized through backpropagation like in the main encoder, the weights of the EMA-encoder are updated as an exponential moving average of the main encoder's weights. This has been shown to prevent collapse during training \cite{lecun2022path,grill2020bootstrap}.
This component takes as input the set of masked patches $\mathcal{P}_{\mathcal{M}}$ and produces representations $t_{\mathcal{M}} = E_{\bar{\theta}}(\mathcal{P}_{\mathcal{M}})$, which serve as the target for the predictor.
Additional details are provided in Appendix.

The learning task consists of predicting the encoding of the masked tokens from those of the non-masked tokens in a latent space, specifically by minimizing
\begin{center}
  $\mathcal{L} = \frac{1}{\mid \mathcal{M} \mid} \sum_{i \in \mathcal{M}} \mid\mid z'_i - t_i \mid\mid_1 = \frac{1}{\mid \mathcal{M} \mid} \sum_{i \in \mathcal{M}} \mid\mid  P_{\beta}(E_{\theta}(\mathcal{P}_{\mathcal{N}}))_i -  E_{\bar{\theta}}(p_i) \mid\mid_1 .$
\end{center}

\section{Experimental Setup}

We first pretrain TS-JEPA using all available training samples for each dataset. Subsequently, we evaluate the learned representation using a frozen evaluation protocol.
Specifically, we keep the pre-trained encoder fixed and train a small classification/regression heads for the downstream tasks.
This allows us to assess the quality and transferability of the learned representations.
For simplicity, we maintain the same architecture for both our encoder and predictor, i.e. a transformer with $2$ attention heads and an embedding dimension of $128$.
All the models have been trained using the AdamW optimizer~\cite{loshchilov2017decoupled}, and the learning rates have been tuned to maximize the performance of each method.
We report the specific details in the Appendix.

\textbf{Baselines.} We compare the performance of TS-JEPA against three alternative methods: (i) TS2Vec~\cite{yue2022ts2vec}, based on contrastive learning, (ii) MAE~\cite{he2022masked}, which employs a mask-and-reconstruct paradigm in the input space and (iii) an autoregressive approach, which predicts future values based on past observations.
We always use the same architecture as the one used in TS-JEPA's encoder (in terms of dimensions and number of attention heads) to ensure the fairness of the comparison. To assess the effectiveness of the encoder's pretraining, we also report results obtained by training only the classification/regression head on a frozen, randomly initialized (non-pretrained) encoder. This comparison helps determine whether the encoder has truly captured the underlying patterns of the dataset or if the task is so simple that a linear layer alone can achieve good performance. We additionally consider the fully supervised case, where a transformer and a CNN are trained end-to-end on the downstream tasks. These supervised baselines provide an upper bound to gauge how close our approach can get to fully supervised learning.

\textbf{Datasets.} For classification, we use FordA, FordB~\cite{dau2019ucr}, FaultDetectionA and FaultDetectionB~\cite{lessmeier2016condition}, all of which comprise outputs from various sensors. Additionally, we include ECG500 ~\cite{dau2019ucr}, consisting of single-sensor electrocardiogram recordings.
For forecasting tasks, we employ the Weather dataset~\cite{MPGWeatherData}, which contains recordings of diverse meteorological indicators. We also consider ETT-Small~\cite{Informer}, representing electricity transformer temperature data, and the Electricity dataset~\cite{electricity-load-diagrams}, which measures electric power consumption in a single household.

\textbf{Evaluation tasks.} We focus on classification and forecasting. For classification, we consider two settings. The first involves using the same dataset for both pre-training and evaluation, while the second assesses the transferability of the pretrained model to similar datasets with similar tasks. Specifically, we pretrain on FordA (resp. FaultDetectionA) and evaluate on FordB (resp. FaultDetectionB).
We evaluate the performance on the forecasting task in two ways. We first focus on short-term forecasting, where, given the immediate context, we predict the next patch in the sequence. We then consider longer-term forecasting, where we predict a horizon window by autoregressively rolling out the input. 

\section{Experimental Results}
Table~\ref{tab:classification} presents the average prediction accuracy and standard deviations (over $10$ runs) for the classification downstream task.
Our experiments show that TS-JEPA outperforms both contrastive and autoregressive approaches in the majority of classification tasks, while exhibiting comparable performance to the Masked Auto-Encoder approach.
Notably, the resulting accuracy closely approximates that of a transformer trained in a fully supervised manner.

\begin{wrapfigure}{r}{0.4\textwidth}
    \centering
    \vspace{-4pt}
    \includegraphics[width=\linewidth]{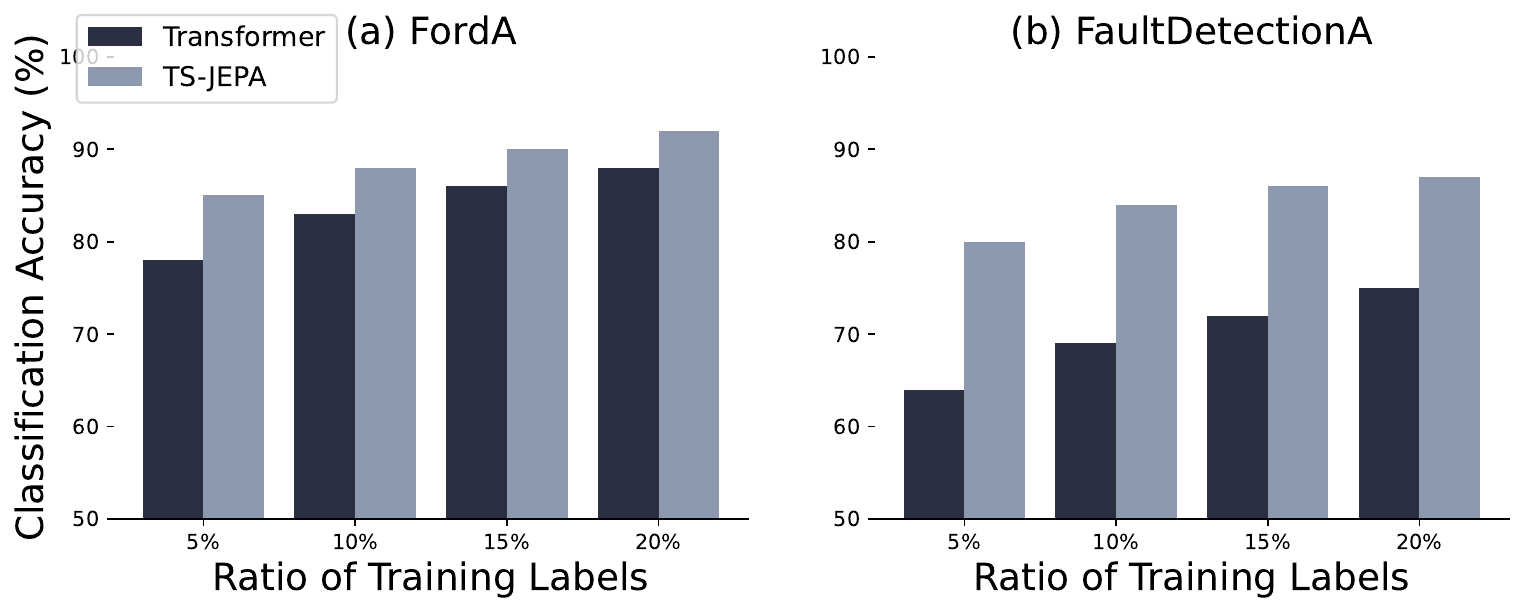}
    \caption{\small Performance of TS-JEPA against a full-supervised Transformer when subject to less training labels on the FordA dataset (a) and FaultDetectionA dataset (b).}
    \label{fig:finetune_result}
    \vspace{-4pt}
\end{wrapfigure}

To further validate the efficacy of our TS-JEPA approach, we conduct an additional evaluation in the context of fine-tuning with limited labeled data.
In this experiment, we mimic a realistic scenario in which supervised labels are sparse. Specifically we only consider a fraction of labels from our labeled data, with varying size from 5\% to 20\% of the total labels and treat the rest of the dataset as unlabeled samples used for pretraining.
Figure~\ref{fig:finetune_result} illustrates the performance of TS-JEPA compared to a fully-supervised transformer model across different amounts of available labeled data.
TS-JEPA shows superior sample efficiency, achieving higher performance with fewer labeled examples. As expected, we observe that the performance gap between TS-JEPA and fully supervised method narrows as the amount of labeled data increases, with the gap shrinking with more labels.

\begin{table}[t]
  \centering
  \caption{Downstream Classification accuracy ($\pm$ standard deviation) using real-world datasets.}
  \renewcommand{\arraystretch}{1.15}
  \resizebox{\textwidth}{!}{%
    \begin{tabular}{lcc|cccccc}
      \toprule
      \multirow{2}{*}{\textbf{Method}} & \multicolumn{2}{c|}{\textbf{Fully Supervised}} & \multicolumn{6}{c}{\textbf{Pre-trained}}                                                                                                                                                                                                                                           \\ \cline{2-9}
                                       & \textbf{CNN}                                   & \textbf{Transformers}                    & \textbf{Trained On} & \textbf{\begin{tabular}[c]{@{}c@{}}Classification \\ Head\end{tabular}} & \textbf{\begin{tabular}[c]{@{}c@{}}TS2Vec \\ (Contrastive)\end{tabular}} & \textbf{MAE}   & \textbf{Auto-Regressive} & \textbf{TS-JEPA} \\ \midrule
      FordA                            & 86.8 $\pm$ 0.4                                 & 91.8 $\pm$ 0.5                           & $\checkmark$        & 46.3 $\pm$ 0.2                                                          & 86.4 $\pm$ 0.2                                                           & 85.1 $\pm$ 0.6 & 69.6 $\pm$ 0.4           & 91.5 $\pm$ 0.1   \\
      FordB                            & 70.5 $\pm$ 0.8                                 & 74.8 $\pm$ 1.1                           & $\times$            & 51.2 $\pm$ 0.6                                                          & 72.4 $\pm$ 0.7                                                           & 59.6 $\pm$ 0.5 & 61.9 $\pm$ 0.3           & 73.8 $\pm$ 0.3   \\ \hline
      FaultDetectionA                  & 98.4 $\pm$ 0.3                                 & 91.8 $\pm$ 0.8                           & $\checkmark$        & 54.3 $\pm$ 2.1                                                          & 83.9 $\pm$ 0.4                                                           & 90.4 $\pm$ 0.3 & 81.6 $\pm$ 0.2           & 85.8 $\pm$ 0.1   \\
      FaultDetectionB                  & 63.9 $\pm$ 1.2                                 & 54.3 $\pm$ 0.4                           & $\times$            & 40.2 $\pm$ 3.7                                                          & 53.9 $\pm$ 0.6                                                           & 54.3 $\pm$ 0.4 & 51.4 $\pm$ 0.7           & 50.6 $\pm$ 1.3   \\ \hline
      ECG5000                          & 87.3 $\pm$ 0.6                                 & 89.9 $\pm$ 1.8                           & $\checkmark$        & 58.4 $\pm$ 0.1                                                          & 86.9 $\pm$ 0.3                                                           & 91.6 $\pm$ 0.7 & 87.5 $\pm$ 0.2           & 89.5 $\pm$ 0.1   \\ \bottomrule
    \end{tabular}%
  }
  \label{tab:classification}
    \vspace{-10pt}
\end{table}

\begin{figure}[h]
  \centering
  \begin{minipage}{0.47\textwidth}
    \centering
    \captionof{table}{MSE and MAE of short-term forecasting.}
    \vspace{0.3em}
    \resizebox{\textwidth}{!}{%
      \begin{tabular}{lclclcl}

        \toprule
        \multirow{2}{*}{Dataset} & \multicolumn{2}{c}{ETT-Small} & \multicolumn{2}{c}{Weather} & \multicolumn{2}{c}{Electricity}                                                             \\ 
                                 & MSE                           & \multicolumn{1}{c}{MAE}     & MSE                             & \multicolumn{1}{c}{MAE} & MSE   & \multicolumn{1}{c}{MAE} \\ \midrule
        Auto-regressive          & 0.009                         & 0.083                       & 0.022                           & 0.108                   & 0.010 & 0.076                   \\
        JEPA                     & 0.017                         & 0.110                       & 0.015                           & 0.109                   & 0.014 & 0.086                   \\ \bottomrule
      \end{tabular}%
    }
    \label{tab:forecasting}
  \end{minipage}
  \hfill
  \begin{minipage}{0.47\textwidth}
    \centering
    \includegraphics[width=\linewidth]{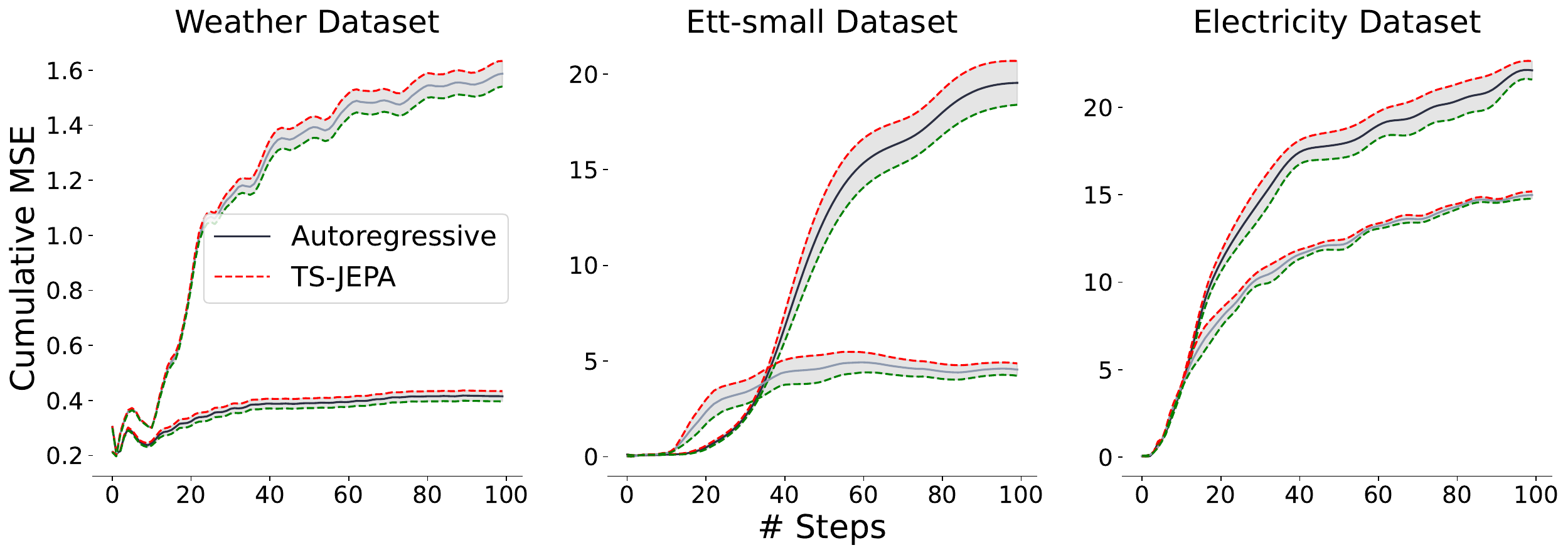} 
    \caption{\small Comparison of the Cumulative Mean Square Error on the long-term forecasting task.}
    \label{fig:forecasting}
  \end{minipage}%
\end{figure}

Table~\ref{tab:forecasting} displays the average MSE and MAE for the short-term forecasting strategy (confidence level provided in the Appendix), while Figure~\ref{fig:forecasting} illustrates the cumulated MSE for long-term forecasting. As anticipated, the autoregressive approach outperforms TS-JEPA in short-term forecasting tasks, which is consistent with the training paradigm of these models.
In long-term forecasting, both approaches exhibit the inherent uncertainty amplification effect characteristic of the roll-out strategy.
However, TS-JEPA demonstrates superior performance to autoregressive strategies in two out of three datasets (i.e. ETT and Electricity), suggesting an enhanced stability.

Overall, these results show promising performance for TS-JEPA in both classification and forecasting tasks.
While the method does not consistently outperform all baselines on every dataset and task, by maintaining competitive forecasting capabilities and outperforming on classification, it offers a compelling trade-off between the two tasks for using a single architecture.

\section{Conclusion}
We introduce TS-JEPA, an adaptation of the JEPA Architecture tailored specifically for self-supervised learning in time series analysis.
The experimental results demonstrate that TS-JEPA achieves a great balance between performance in classification and forecasting downstream tasks.
This balanced capability sets TS-JEPA apart from current state-of-the-art methods, particularly the widely-used autoregressive approach, which often excels in one task (e.g. forecasting) at the expense of the other.
The versatility of TS-JEPA makes it a promising candidate for developing adaptable foundation models for time series analysis. Next steps will include the  exploration of scaling strategies for TS-JEPA, with the ultimate goal of establishing a new paradigm for time series foundation models.
\section*{Acknowledgements}
We thank Alberto Bietti,  Francois Lanusse, Rudy Morel for various discussions. 
The computations in this work were run at facilities at the Flatiron Institute, a division of the Simons Foundation, and we are thankful to the Scientific Computing Core and Shirley Ho for their support.

\bibliographystyle{plain}
\bibliography{references}

\newpage

\begin{center}
  \centering{\LARGE\bf Appendix: Joint Embedding go Temporal.\par}
\end{center}

\setcounter{section}{0}
\section{Pre-training Details}

\textbf{Architecture.} Our encoder and decoder share an identical architecture: a transformer with 2 attention heads and an embedding dimension of 128.
While we have not conducted an ablation study on this configuration, it has proven sufficient to yield satisfactory results in downstream tasks.
We defer a more thorough investigation of architectural choices to future work.
To ensure a fair comparison, all considered baselines utilize the same architecture as our TS-JEPA, allowing us to isolate the effects of the self-supervised task.
Across all datasets and tasks, we segment each time series into 10 patches and employ a batch size of 32.
For TS-JEPA, we apply a masking ratio of $70\%$, while for MAE, we use $75\%$, consistent with the implementation in~\cite{liu2023ssl}.

\begin{figure}[h]
  \centering
  \includegraphics[width=0.95\linewidth]{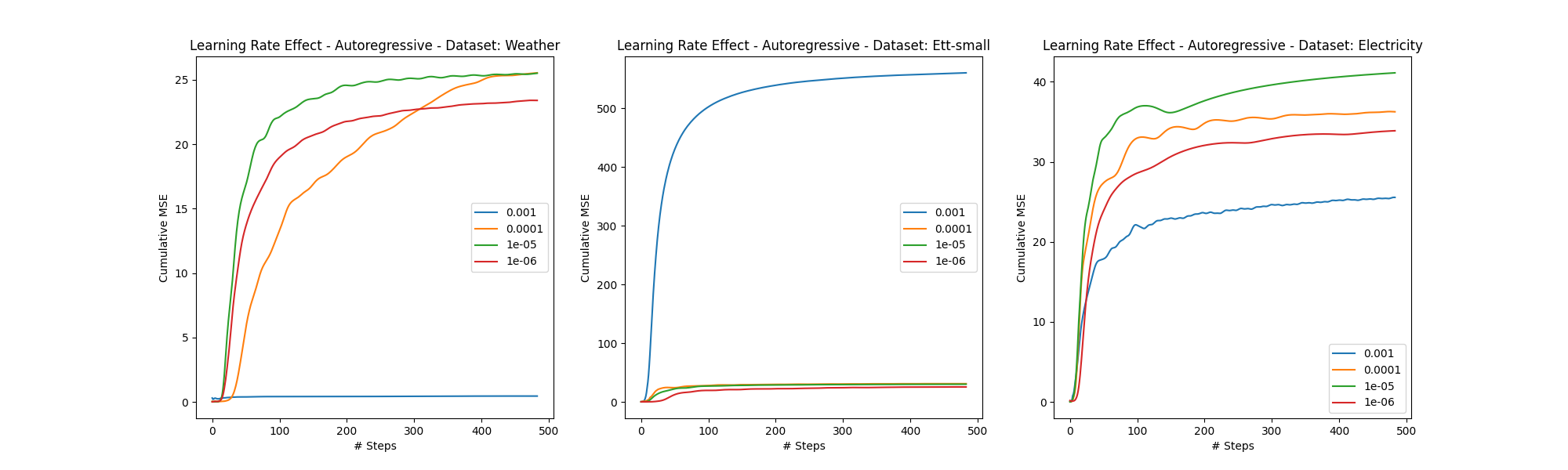}
  \caption{Effect of Learning Rate on the long-term forecasting for autoregressive models.}
  \label{fig:learning_rate_effect_forecasting_autoregressive}
\end{figure}

\begin{figure}[h]
  \centering
  \includegraphics[width=0.95\linewidth]{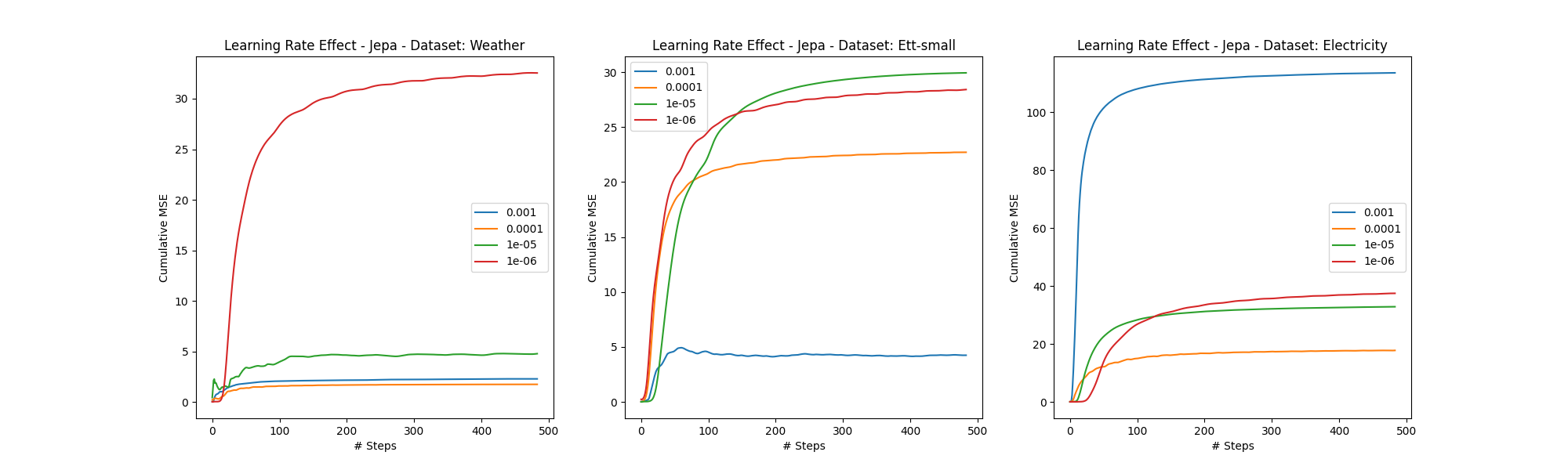}
  \caption{Effect of Learning Rate on the long-term forecasting for JEPA.}
  \label{fig:learning_rate_effect_forecasting_jepa}
\end{figure}

\textbf{Optimization.} We train all models using the AdamW optimizer~\cite{loshchilov2017decoupled}, with learning rates fine-tuned to maximize the performance of each method. Specifically, we explore learning rates within the range [1e-03, 1e-04, 1e-05, 1e-06]. Table \ref{tab:learning_rate_effect} and Figure \ref{fig:learning_rate_effect_forecasting_autoregressive} and \ref{fig:learning_rate_effect_forecasting_jepa} illustrate the impact of learning rate on both long-term and short-term forecasting downstream tasks for both Autogressive approach and TS-JEPA.

\textbf{On the EMA-Encoder} The use of two instances of the same encoder has been shown to lead to a representation collapse, where the encoder learns the trivial solution of a constant output regardless of the input. \cite{lecun2022path}. Typically, this solution minimizes the reconstruction loss but fails to capture any meaningful information about the data. By using the EMA-encoder, we introduce a slowly moving target for the predictor, helping to stabilize training and encourages the model to learn more robust and meaningful representations.
The use of an EMA-encoder has been validated in previous work, including the Bootstrap Your Own Latent (BYOL) \cite{grill2020bootstrap} method for self-supervised learning, and in JEPA architectures applied to images \cite{assran2023self} and videos \cite{bardes2024vjepa}. For our implementation we have used set the update parameter $m=0.998$, which we have seen to emperically give good results while avoid the collapse.

\begin{table}[h]
  \centering
  \caption{Effect of Learning Rate on both JEPA and Autoregressive approach.}
  \renewcommand{\arraystretch}{1.2}
  \resizebox{\textwidth}{!}{%
    \begin{tabular}{llclclclcl}
      \hline
      \multirow{2}{*}{\textbf{Dataset}}     & \textbf{Learning Rate:}   & \multicolumn{2}{c}{$1e-03$} & \multicolumn{2}{c}{$1e-04$} & \multicolumn{2}{c}{$1e-05$} & \multicolumn{2}{c}{$1e-06$}                                                                                                             \\ \cline{2-10}
                                   & \textbf{Metric:}          & MSE                         & \multicolumn{1}{c}{MAE}     & MSE                         & \multicolumn{1}{c}{MAE}     & MSE                       & \multicolumn{1}{c}{MAE} & MSE                       & \multicolumn{1}{c}{MAE} \\ \hline
      \multirow{2}{*}{ETT-Small}   & Auto-regressive & 0.035                       & 0.123                       & \multicolumn{1}{l}{0.028}   & 0.113                       & 0.009                     & 0.083                   & 0.042                     & 0.152                   \\ \cline{2-10}
                                   & JEPA            & 0.031                       & 0.133                       & 0.031                       & 0.114                       & 0.017                     & 0.110                   & 0.019                     & 0.114                   \\ \hline
      \multirow{2}{*}{Weather}     & Auto-regressive & 0.054                       & 0.153                       & 0.039                       & 0.160                       & 0.022                     & 0.108                   & 0.030                     & 0.124                   \\ \cline{2-10}
                                   & JEPA            & 0.017                       & 0.101                       & 0.053                       & 0.172                       & 0.037                     & 0.147                   & 0.015                     & 0.109                   \\ \hline
      \multirow{2}{*}{Electricity} & Auto-regressive & \multicolumn{1}{l}{0.013}   & 0.089                       & \multicolumn{1}{l}{0.022}   & 0.114                       & \multicolumn{1}{l}{0.010} & 0.076                   & \multicolumn{1}{l}{0.048} & 0.145                   \\ \cline{2-10}
                                   & JEPA            & \multicolumn{1}{l}{0.014}   & 0.086                       & \multicolumn{1}{l}{0.058}   & 0.198                       & \multicolumn{1}{l}{0.022} & 0.121                   & \multicolumn{1}{l}{0.026} & 0.140                   \\ \hline
    \end{tabular}%
  \label{tab:learning_rate_effect}
    
  }
\end{table}

\begin{figure}[h]
  \centering
  \includegraphics[width=0.95\linewidth]{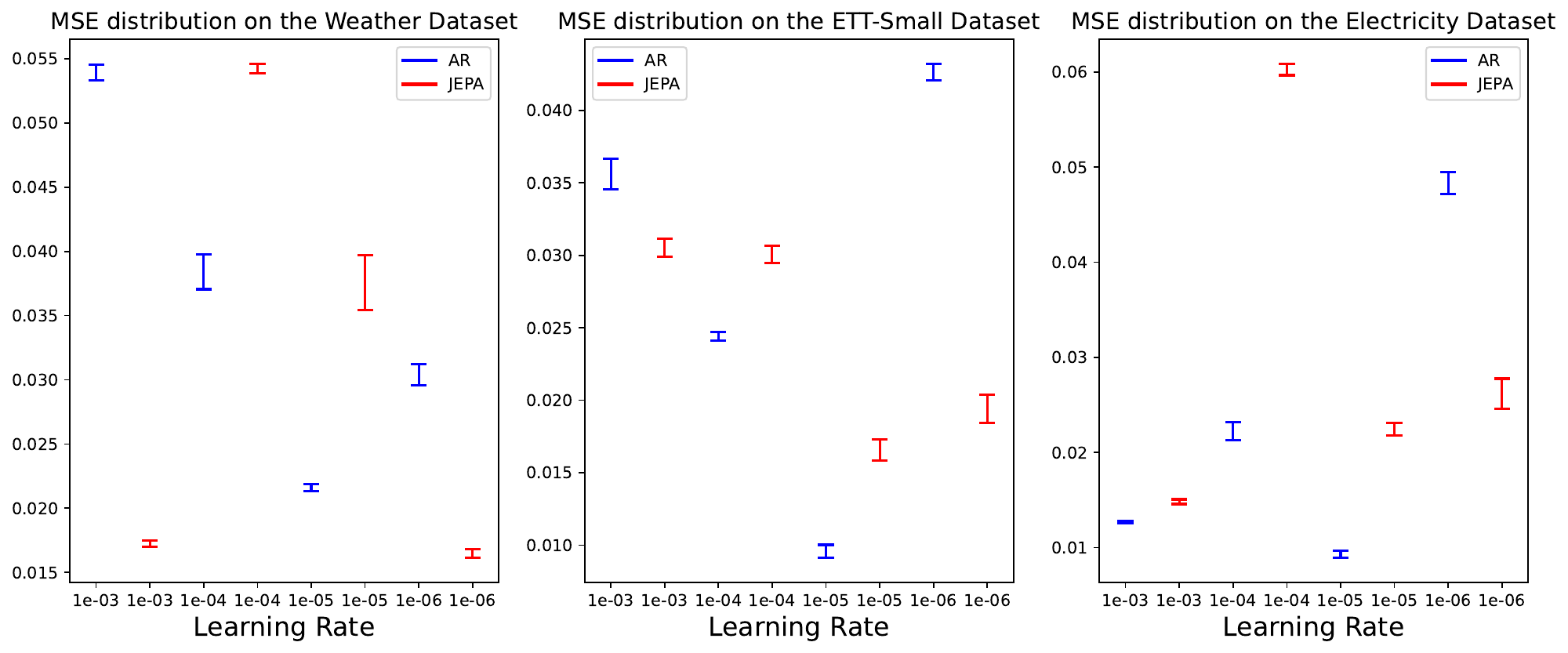}
  \caption{MSE distributions for each experiment.}
  \label{fig:MSE_distribution}
\end{figure}

\section{Datasets and Implementation Details}

\textbf{Implementation details.} The experimental implementation of TS-JEPA is available on github\footnote{\href{https://github.com/Sennadir/TS\_JEPA}{https://github.com/Sennadir/TS\_JEPA}}. It is built using the open-source PyTorch library. We use the official, publicly available implementation of TS2Vec \cite{yue2022ts2vec} and the MAE implementation provided by \cite{liu2023ssl}. All experiments were conducted on an NVIDIA V100 GPU.

\begin{table}[h]
\caption{Statistics of the classification datasets used in our experiments.}
\label{tab:data_statistics}
\vskip 0.15in
\begin{center}
\begin{small}
\begin{sc}
\begin{tabular}{lcccc}
\toprule
Dataset & \#Training Point & \#Test Points & \#Length & \#Classes \\
\midrule
FordA    & 3601 & 1320 & 500 & 2 \\
FordB    & 3636 & 810 & 500 & 2 \\
FaultDetectionA    & 10912 & 2728 & 5120 & 3 \\
FaultDetectionB    & 10912 & 2728 & 5120 & 3 \\
ECG500    & 500	& 4500 & 140 & 5 \\

\bottomrule
\end{tabular}
\end{sc}
\end{small}
\end{center}
\vskip -0.1in
\end{table}

\end{document}